\documentclass{new_tlp}

\usepackage{graphicx,url}

\newcommand\bcmdtab{\noindent\bgroup\tabcolsep=0pt%
  \begin{tabular}{@{}p{10pc}@{}p{20pc}@{}}}
\newcommand\ecmdtab{\end{tabular}\egroup}

  \title[Tabling for Petrobras Planning]
        {Using Tabled Logic Programming to Solve the Petrobras Planning Problem}

  \author[R. Bart\'{a}k and N.-F. Zhou]
         {ROMAN BART\'{A}K\\
         Charles University in Prague, Faculty of Mathematics and Physics, Prague, Czech Republic\\
         \email{bartak@ktiml.mff.cuni.cz}
         \and NENG-FA ZHOU\\
         The City University of New York, Brooklyn College, New York, USA\\
         \email{nzhou@acm.org}}

\jdate{February 2014}
\pubyear{2014}
\pagerange{\pageref{firstpage}--\pageref{lastpage}}
\doi{XXX}

\begin{document}

\label{firstpage}

\maketitle

  \begin{abstract}
Tabling has been used for some time to improve efficiency of Prolog programs by memorizing answered queries. The same idea can be naturally used to memorize visited states during search for planning. In this paper we present a planner developed in the Picat language to solve the Petrobras planning problem. Picat is a novel Prolog-like language that provides pattern matching, deterministic and non-deterministic rules, and tabling as its core modelling and solving features. We demonstrate these capabilities using the Petrobras problem, where the goal is to plan transport of cargo items from ports to platforms using vessels with limited capacity. Monte Carlo Tree Search has been so far the best technique to tackle this problem and we will show that by using tabling we can achieve much better runtime efficiency and better plan quality.
  \end{abstract}

  \begin{keywords}
    tabling, resource-bounded search, symmetry breaking, planning, logistics
  \end{keywords}


\section{Introduction}

Prolog and other logic programming languages are usually not assumed as the first-choice programming language for developing real-life planning applications. In this paper we shall demonstrate that Picat, a novel Prolog-like language, provides strong support to solve hard and realistic planning problems. In particular, we will propose a method to solve the Petrobras logistic problem introduced in the \emph{Fourth International Competition on Knowledge Engineering for Planning and Scheduling} (ICKEPS 2012). Picat allowed us to develop a declarative model of the problem. We exploited Picat tabling as a mechanism to help the classical depth-first search to resolve non-deterministic decisions. We will experimentally show that the proposed model beats the existing techniques both in runtime and in generating plans with lower fuel consumption.

The paper is organized as follows. We will first describe the Petrobras planning problem and survey existing techniques applied to this problem. After that we will introduce our solving approach which is based on decomposition of the problem and on tabling-friendly representation of search states. This allowed us to exploit tabling to remove re-exploration of already visited search states. Finally, we will present experimental comparison of our method with existing techniques.

\section{Petrobras Problem}

The Petrobras problem has been introduced in the \emph{Fourth International Competition on Knowledge Engineering for Planning and Scheduling} (ICKEPS 2012) as one of real-life challenge problems. The major task of the competition was to model the problems formally as well as to solve them. In this section we will first give some details of the Petrobras problem and then we will survey the techniques that have been applied to solve the problem.

\subsection{Problem Specification}
The Petrobras problem is a logistic problem of planning cargo deliveries from ports to oil platforms. This problem is motivated by a real-life problem of the Brazilian oil company Petrobras. A complete description of the problem including example data can be found in \cite{petrobras}.

\begin{figure}
 \begin{center}
   \includegraphics[width=0.7\textwidth]{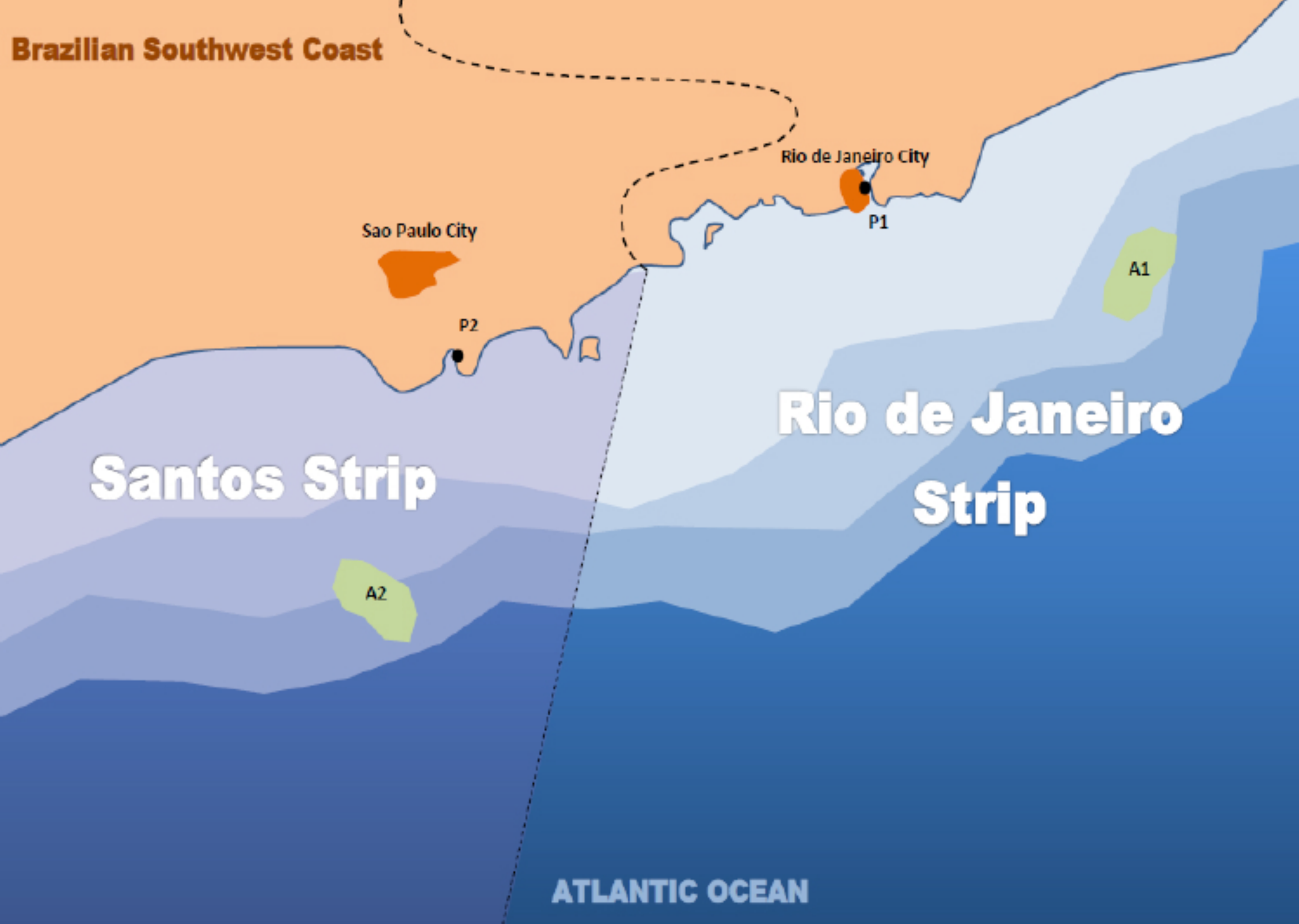}
 \end{center}
    \caption{Position of ports and waiting areas in the Petrobras problem \protect\cite{petrobras}.}
    \label{PetroArea}
 \end{figure}

The problem can be described as follows. There are several identical vessels with limited cargo capacity and initial fuel levels. At the beginning each vessel is in one of two waiting areas (marked A1 and A2 in Figure~\ref{PetroArea}). A set of cargo items is prepared in two ports (marked P1 and P2 in Figure~\ref{PetroArea}). Each cargo item has a specific weight and a destination where the item should be delivered. Vessels navigate to ports where they can load cargo items while respecting the vessels' capacity limits. Then the vessels navigate to platforms where they unload the cargo. After entering a port or a platform (and before loading and unloading), each vessel must dock. The vessel must undock before leaving the port and the platform. At most two vessels can be docked simultaneously in ports and at most one vessel can be docked at a platform. If a port or a platform is occupied then the next vessel must wait before docking. Docking and undocking operations take a constant time while the durations of loading and unloading operations depend linearly on the weight of cargo. Vessels consume fuel when they navigate between waiting areas, ports, and platforms. The quantity of consumed fuel and the time necessary to travel depend linearly on the distance between the locations and also on the load of the vessel. Only two levels of load, loaded and empty, are distinguished. Vessels have limited-capacity fuel tanks; the fuel tanks can be refilled in ports and in some platforms. It is important to ensure that at any time, each vessel has enough fuel to go to the nearest refueling station. At the end, the vessels are supposed to be back in one of the waiting areas.

Summarizing the above text, there are six operations, \emph{navigate, dock, undock, load, unload}, and \emph{refuel}, that each vessel can do. Only refueling can run in parallel with loading or unloading, while any other pair of operations for a single vessel cannot overlap in time. The vessel must be docked before loading, unloading, and refueling, and it must be undocked before navigating. We are given the initial locations and fuel levels of vessels and the initial location, weight, and destination of each cargo item. The task is to plan operations for vessels in such a way that all cargo items are delivered. 

The problem formulation contains several metrics, namely the total consumed fuel, the number of vessels used to solve the problem, the size of waiting queues, the total amount of time (makespan), and the docking cost. All these metrics are supposed to be minimized.

\subsection{Existing Solving Approaches}
Several techniques have been tried to tackle the Petrobras challenge. The first three approaches, namely classical planning, temporal planning, and Monte Carlo Tree Search were introduced in \cite{toropila2012}, while the forth approach based on finite state automata and macro actions is described in \cite{bartak2013}.

The \emph{classical planning} approach modeled the problem in PDDL 3.0 \cite{pddl3} with numerical fluents describing the restricted resource capacities (fuel tank, cargo capacity). This model used actions as specified in the problem formulation, namely: \emph{navigate-empty-vessel, navigate-nonempty-vessel, load-cargo, unload-cargo, refuel-vessel-platform, refuel-vessel-port, dock-vessel, undock-vessel}. SGPlan 6.0 \cite{sgplan} was used to solve the problem while optimizing fuel consumption. Action durations were added to the solution in the post-processing stage.

The \emph{temporal planning} approach modeled the problem in PDDL 3.1 \cite{pddl31} with fluents and durative actions. Basically the same set of actions as in the classical planning approach  was used with added durations. This model supports concurrency of actions directly. The Filuta planner \cite{dvorak2010} was used to solve the problem while optimizing makespan. Filuta uses ad-hoc solvers to handle resources, namely unary, consumable, and reservoir resources are supported.

The third approach exploited \emph{Monte Carlo Tree Search} (MCTS) techniques that become recently popular in computer Go. To allow using MCTS, a different action model was applied to obtain finite plans in search branches. This model is based on four abstract actions: Load, Unload, Refuel, GoToWaitingArea. These actions describe ``intentions'' and they are decomposed to primitive actions based on situation (state).
The MCTS solver used a specific objective function combining several metrics: $usedFuel + 10*numActions + 5*makespan$.

Among the above three approaches, the MCTS was the clear winner as it generated plans with both smallest makespan and smallest total fuel consumption. The Filuta planner was also able to solve all benchmark problems (up to 10 vessels and up to 15 cargo items) but the generated plans were consistently worse than the plans obtained by MCTS (about 30\% longer makespan and 130 \% more fuel). This motivated another approach whose goal was to explain the good performance of MCTS. This last approach was again based on grouping sequences of actions into macro-actions similar to those in the MCTS method. The allowed sequences of macro-actions were described by a finite state automaton \cite{bartak2013} which guides the planner. Straightforward left-to-right integrated planning and scheduling was used with simple heuristic estimating makespan, which was the primary objective. This solver was implemented in B-Prolog and it was the first attempt to exploit a tabling mechanism in the Petrobras problem. Unfortunately, due to involvement of time variables, the number of states was too large for tabling when more than six cargo items were assumed. This approach generated plans with makespan smaller or comparable to MCTS, but with the total fuel consumption closer to the Filuta planner (see Figure~\ref{GraphFuel}). 

\section{Motivation for Tabling in Planning}

Though the first logic programming language PLANNER \cite{planner} was designed as ``a language for proving theorems and manipulating models in a robot'' and planning has been an important problem domain for Prolog \cite{warplan}, there has been little success in applying Prolog to planning. One of the reasons is the built-in computation mechanism based on depth-first search where all information collected in one search branch is forgotten when backtracking and exploring a parallel search branch. Hence straightforward Prolog-based planners are prone to revisiting already explored states and to looping. \emph{Tabling} \cite{tabling} is a mechanism that can help there without sacrificing the declarative nature of Prolog programs.

Though planning techniques made a great leap in recent years, there are still many problems where classical planning approaches fail. One of such problems is Sokoban that has been adopted for ASP and IPC competitions \cite{asp-soko}. The early B-Prolog program with tabling showed the capabilities of tabling by providing a nice declarative model of the Sokoban problem that lead to a very good performance \cite{sokoban}. However, the direct motivation for our research of exploiting tabling in the Petrobras domain was the famous \emph{Nomystery} domain used in IPC 2011 and ASP 2013 \cite{nomystery}. Nomystery is a transportation planning problem that can be described as follows. Given a weighted directed graph modeling a network of locations, a truck that is located initially at a location, and a set of packages each of which has an initial location and a destination, the objective of the problem is to find a plan for the truck to transport the packages from their initial locations to their destinations. The truck has some initial fuel and each transportation action (moving from location to another location) consumes some fuel defined by the weight of the edge in the graph. An optimization version of the problem can be defined as finding a plan with the minimal total fuel consumption.

We solved the Nomystery problem in B-Prolog \cite{bprolog} and later in Picat \cite{picat} in a straightforward way as finding a path in the state space. We modeled the state as a current location of truck, its fuel level, a set of loaded cargo, and a set of cargoes to be delivered. Three types of state transitions were assumed, namely moving between two connected locations, loading cargo, and unloading cargo. A non-negative fuel level was required in each state. The only non-determinism to be explored by the search algorithm was moving between locations as we handled loading and unloading cargo deterministically using the following rules:
\begin{itemize}
\item if the truck is in a location where some cargo to deliver is placed then the cargo is always loaded to the truck, and
\item if the truck is at a location where some loaded cargo should be delivered then the cargo is always unloaded from the truck.
\end{itemize}
This approach was possible as there is no capacity limit for cargo in the truck and hence we can load as many cargo items as we need, move them together, and unload them only in their destinations. This simple declarative model in Picat performed surprisingly well and it solved all of the Nomystery instances (in less than one minute per instance) used in IPC 2011, including the hardest instance which was not solved by any of the participating planners.

The Nomystery domain is very close to the Petrobras problem which motivated us to explore a similar modelling technique to solve the Petrobras problem. The previous attempt using B-Prolog \cite{bartak2013} was promising but the results regarding total fuel consumption were not as good as those from the MCTS approach. Hence we focus on minimizing the fuel consumption which is also the major metric in the Nomystery domain. There are following major differences of the Petrobras problem from the Nomystery domain (assuming vessels as straightforward counterparts for trucks):
\begin{itemize}
\item the capacity of a vessel to transport cargo is limited (hence it is not possible to load all cargo to a single vessel),
\item the fuel capacity of a vessel is limited (hence refueling may be necessary to visit required destinations),
\item there are more than one vessel.
\end{itemize}

\section{A Solving Approach}

To solve the Petrobras problem we decomposed the problem into sub-problems solved separately. If the decomposition is done right then it significantly increases efficiency of the solver as the sub-problems can be solved independently. On the other hand, using decomposition may hamper solution quality as the combination of optimal solutions of sub-problems may not be the optimal solution of the original problem. We focus on optimizing fuel consumption in the Petrobras problem and we will discuss how our decomposition decisions influence feasibility and optimality of solutions.

We first decomposed the problem into planning and scheduling parts. The \emph{planning task} consists of deciding which actions are necessary to deliver all cargo. Fuel consumption and limited fuel and cargo capacities of vessels are assumed during planning, while actions' durations and port and platform docking capacities are left to scheduling. The \emph{scheduling task} then consists of allocating actions to particular times. One should realize that a feasible plan can always be scheduled. The plan describes causal relations between the actions so we can allocate the actions to earliest possible start times while respecting these causal (precedence) relations. The port and platform docking capacity constraints can also be easily satisfied as the vessel can wait when the port or platform is occupied by another vessel. Hence for any feasible plan we can find a feasible schedule.

Fuel consumption depends only on the plan and it is independent of particular time allocation (fuel is consumed by the navigation actions and these actions are decided during the planning stage). Hence optimizing fuel consumption can be done during planning without being influenced by the scheduling decisions. In the rest of the paper, we will focus on the planning task with the main objective to minimize consumed fuel.

Splitting the Petrobras problem to planning and scheduling stages has already been applied in the classical planning approach that we described above \cite{toropila2012}. That approach was not successful because the planner was not able to generate the plans for larger problems. Hence proposing an efficient planner is an interesting task itself. The tabled approach to solve the Nomystery domain demonstrated that using tabling is a promising direction to solve logistic problems.

The classical planning task is basically about finding a sequence of transitions in the state space. Using a straightforward model with world states and transitions defined by the primitive actions would not work (this is what the classical planning approach tried). Tabling itself converts tree search to graph search so it can only remove repeated exploration of already visited states. However, if there are many reachable states then tabling fills up memory very fast. Hence, we decided for further problem decomposition with a careful definition of search states and transitions to remove symmetries. 

\subsection{Symmetries and Representation of States}
There are many symmetries in the Petrobras problem. First, all vessels in the problem are identical, meaning they have exactly the same weight capacity and fuel capacity limits and the same speed and fuel consumption that only depends on whether any cargo is loaded or not. If two vessels with identical fuel levels are located at the same place then these vessels are indistinguishable in the plan. Hence we identify vessels by location and fuel level (see below) so we can easily recognize such symmetries and select the first vessel from the set of identical vessels deterministically during planning. There is a similar symmetry of cargo items. Two cargo items located in the same port with identical weights and destinations are indistinguishable. Though this situation is rare, we handle it in the procedure for selecting cargo for transport (see below).

Based on above observations, we represent the main search states as positions of cargo items and vessels. We completely omit identifications of vessels and cargo items in this representation. A vessel is identified by its location and a fuel level, while the cargo item is identified by its origin, destination and weight. We group the vessels based on their location and similarly we group the cargo items first based on their origin and then by their destination. In particular, we represent the set of cargo items using a list of lists:
\begin{verbatim}
     [[OriginLoc, [DestinationLoc, Weight1,Weight2,...]], ...]
\end{verbatim}
where the list of cargo weights is sorted downward to remove symmetrical list representations of the same set. Similarly, we represent the set of vessels as the following list of lists:
\begin{verbatim}
     [[Location, FuelLevel1, FuelLevel2,...], ...]
\end{verbatim}
The list of fuel levels is sorted upward to remove symmetries. For example, the initial state of the Petrobras problem \cite{petrobras} is represented as follows:
\begin{verbatim}
   Cargo = [[p1,[f1,20],[f2,30,15],[f3,10],[f4,15],[f5,25],[f6,5]],
            [p2,[f1,40],[f2,30],[f3,20],[g1,20,15],[g2,30,20],[g4,8]]]
   Vessels = [[a1,400,400,400,400,400,400],
              [a2,400,400,400,400]]
\end{verbatim}

\subsection{Representation of Transitions}
In the cargo and vessel representations, we do not model states where cargo is loaded to vessels. Motivated by successful MCTS and B-Prolog approaches, we define the state transition as a sequence of following operations:
\begin{enumerate}
\item (if necessary) move an empty vessel to a port with cargo waiting to delivery,
\item dock the vessel, refuel it (ports are among the refueling stations), load selected cargo, and undock the vessel,
\item as an option, move the vessel to another port with some cargo waiting and repeat step (2) or continue with the next step,
\item transport the loaded cargo to required destinations and finish with an empty vessel either in a port with some cargo waiting or at the waiting area.
\end{enumerate}
After performing the above steps, the selected vessel will be empty again located in a waiting area or at a port with cargo to deliver. Basically, there are following branching decisions in steps (1)-(3):
\begin{enumerate}
\item select a port,
\item select a vessel for delivery,
\item select cargo in the port while respecting capacity limit of the vessel (note that all vessels have the same capacity limit),
\item plan delivery of cargo items.
\end{enumerate}
We explore all ports with some undelivered cargo. For the selected port we deterministically select a vessel closest to that port (if there are more vessels at the closest location then the vessel with the lowest fuel level is selected). Then we explore all subsets of cargo items in the port such that the weight of selected cargo is below the vessel weight limit. From these cargo sets we exclude the sets that can be ``improved'' by adding a cargo item going to a destination that is already included in the set. For example, if the cargo set $Cargo$ contains items going to destinations $\{f1, f3, g2\}$ and there is another cargo item $C$ going to destination $f3$ such that $C$ can be added to $Cargo$ while respecting the weight capacity limit, then the set $Cargo$ is not assumed as a viable set for delivery ($Cargo \cup \{C\}$ can be delivered using the same fuel consumption as $Cargo$, see the explanation below). If cargo in the first port is decided, we use the same process to decide if some cargo will be loaded in another port (now, the vessel is known so we only decide the cargo set and move the vessel to the next port). The exploration of alternatives is done using standard backtracking with tabling.

The Picat encoding of the above decision sequence is given in Figure~\ref{PlannerCode}. Picat uses similar programming principles as Prolog with some extensions such as functions (e.g. \verb(addCargo, addVessel() and possibility to denote deterministic rules (\verb|=>|) with guards describing when the rule can be selected (see the second rule in Figure~\ref{RouteCode}). The keyword \verb|tabling| denotes that the next predicate will be tabled. The '\verb|+|' arguments are input, the '\verb|-|' arguments are output, and the '\verb|min|' argument is being minimized. Briefly speaking, for all input arguments of a query, the tabling mechanism remembers the computed outputs so the next time when a query with the same input is asked, the answer is recovered from memory rather than recomputed. If there is a minimization parameter then the tabling mechanism remembers only those answers with the smallest value of this argument. In our particular example, for each state (cargo items to deliver and vessels' locations) we remember the plan with the smallest fuel consumption. The details of Picat syntax and semantics are described in \cite{picat}.

\begin{figure}
\figrule
\begin{verbatim}
table (+,+, -,min)
plan([], _Vessels, Plan, Fuel) =>
     Plan = [], Fuel = 0.
plan(Cargo, Vessels, Plan, Fuel) =>
     select_port(Cargo, Port, PortCargo, RestCargo),
     select_cargo(PortCargo, Destinations, FreeCap, RestPortCargo),
     select_and_move_vessel(Vessels, Port, FuelLevel1, RestVessels, Plan1, Fuel1),
     load_at_other_ports(RestCargo, Port, FreeCap, FuelLevel1,
     	                   Destinations2, RestCargo2, Port2, FuelLevel2, Plan2, Fuel2),
     path_plan(Port2, FuelLevel2, Destinations ++ Destinations2,
     	         FinalLoc, FinalLevel, Plan3, Fuel3),
     plan(addCargo(RestCargo2, Port, RestPortCargo),
          addVessel(RestVessels, FinalLoc, FinalLevel),
          Plan4, Fuel4),
     Plan = Plan1 ++  $[load(Port),undock(Port)] ++ Plan2 ++ Plan3 ++ Plan4,
     Fuel = Fuel1 + Fuel2 + Fuel3 + Fuel4.
\end{verbatim}
\caption{Code in Picat for the main planning loop.}\label{PlannerCode}
 \figrule
 \end{figure}

It remains to describe how to find a plan to deliver loaded cargo items. First, one should realize that this plan does not depend on cargo weights, but the plan depends on the delivery locations only. This observation is very important because it allows us to define the search space with states identified with the location of vessel, its fuel level, and a set of destinations to visit. We can again use tabling to find a plan with minimal fuel consumption as Figure~\ref{RouteCode} shows. Note that refueling is hidden in navigation between two delivery locations. If there is not enough fuel then a refueling station is visited in the route. Moreover when the vessel visits a platform with a refueling station, we always plan a refueling operation there.
 
\begin{figure}
\figrule
\begin{verbatim}
table (+,+,+, -,-,-,min)
path_plan(Loc, FuelLevel, [], FinalLoc, FinalLevel, Plan, Fuel) =>
     FinalLoc = Loc, FinalLevel = FuelLevel, Plan = [], Fuel = 0.
path_plan(Loc, FuelLevel, ToVisit, FinalLoc, FinalLevel, Plan,Fuel),
     select(Loc, ToVisit, RestToVisit)      % guard when the rule is applicable
 =>
     Plan = $[dock(Loc), unload(Loc), undock(Loc) | RestPlan],
     path_plan(Loc, FuelLevel, RestToVisit, FinalLoc, FinalLevel, RestPlan, Fuel).
path_plan(Loc, FuelLevel, ToVisit, FinalLoc, FinalLevel, Plan,Fuel) =>
     member(NextLoc, ToVisit),
     navigate(Loc,NextLoc, FuelLevel, NextLevel, Plan1, Fuel1),
     path_plan(NextLoc, NextLevel, ToVisit, FinalLoc, FinalLevel, Plan2, Fuel2),
     Plan = Plan1 ++ Plan2,
     Fuel = Fuel1 + Fuel2.
\end{verbatim}
\caption{Code in Picat for the route planning loop.}\label{RouteCode}
 \figrule
 \end{figure}

The global plan is combined from partial plans to deliver sets of cargo items. This way we significantly reduced the search space, but obviously there is no guarantee to find the globally optimal plan. Nevertheless as the experimental evaluation showed, this approach can still find plans better than the existing approaches.

\subsection{Branch and Bound}
One of the disadvantages of pure tabling when applied to optimization problems is exploration of solutions that are known to be worse than a solution found so far. If the solution cost is not decreasing when continuing search, one can use some form of resource-bounded search to remove this deficiency. This is for example the case of path finding with non-negative-cost arcs. Fuel consumption is another example of non-decreasing cost as delivering more cargo items will never require less fuel.

In our approach we suggest using branch-and-bound in the top search procedure  \verb(plan(. The method works as follows. We keep the quantity of already consumed fuel as an extra non-tabled parameter of the \verb(plan( procedure. To estimate the quantity of fuel required to deliver the remaining cargo we can use the procedure \verb(path_plan( applied to a relaxed problem. We relax the weight limit of vessels and we assume that an empty vessel is ready in each port. Using \verb(path_plan( we compute the optimal path to deliver that cargo. We relax the weight capacity limit of the vessel but we assume the fuel tank capacity limit in this computation. If the sum of already consumed fuel and expected fuel exceeds the bound then we cut search and backtrack. This is a form of look-ahead and it is used in all modern search algorithms including the famous A*. The bound is stored on a blackboard as the fuel consumption of the best plan found so far. The initial bound is set to a large-enough number.

\section{Experimental Results}
We first experimentally compare our system with the Filuta planner and the MCTS solver from \cite{toropila2012} and with the B-Prolog system from \cite{bartak2013}. We omit the SGPlan (classical planning) from the comparison as it was not competitive and cannot solve problems with more than six cargo items. We re-use here the results reported in \cite{toropila2012}, where the experiments were run on the Ubuntu Linux machine equipped with Intel Core i7-2600 CPU @ 3.40GHz and 4GB of memory and the planners were allowed to use approximately 10 minutes of runtime. For the B-Prolog implementation we re-use the results reported in \cite{bartak2013}, where the experiments run on the MacOS X 10.7.5 (Lion) machine with 1.8 GHz Intel Core i7 CPU and 4GB of memory and the best results found within one minute of runtime were presented. Our Picat implementation runs under MacOS X 10.9.1 (Mavericks) machine with 1.7 GHz Intel Core i7 CPU and 8GB of memory. 

The original specification of the Petrobras problem \cite{petrobras} contains just one set of real-life motivated data. This set consists of two waiting areas, two ports, ten platforms, ten vessels, and fifteen cargo items. Table~\ref{TablePetro} shows the comparison of major metrics for all tested solvers for the Petrobras problem. The Picat-based planner improved the already very good fuel consumption of the MCTS solution. Moreover, the Picat-based planner proved optimality taking into account the assumptions restricting allowed plans and it run orders of magnitude faster than all other approaches. Makespan is not that good which is caused by using only three vessels (to get good fuel efficiency) while the other approaches used more vessels in parallel. In fact, the Picat-generated plan can use two vessels to decrease the docking cost to 311k while preserving the values of other metrics (a different schedule is used).

\begin{table}
\caption{The results for the Petrobras problem \protect\cite{petrobras}.}\label{TablePetro}
\begin{tabular}{l|r|r|r|r|r}
\hline\hline
System &Fuel Consumption &Vessels &Makespan &Docking Cost &Runtime (ms)\\
\hline
Filuta & 1989 (2.45$\times$) & 5 & 263 & 333 000 & \~{}600 000\\
B-Prolog &1263 (1.56$\times$) & 4 & \textbf{162} & \textbf{311 000} & \~{}60 000\\
MCTS & 887 (1.09$\times$) & 4 & 204 & \textbf{311 000} & \~{}600 000\\
Picat & \textbf{812} (1.00$\times$) & \textbf{3} & 341 & 313 000 & \textbf{813}\\
\hline\hline
\end{tabular}
\end{table}

\begin{figure}
 \begin{center}
   \includegraphics[width=0.7\textwidth]{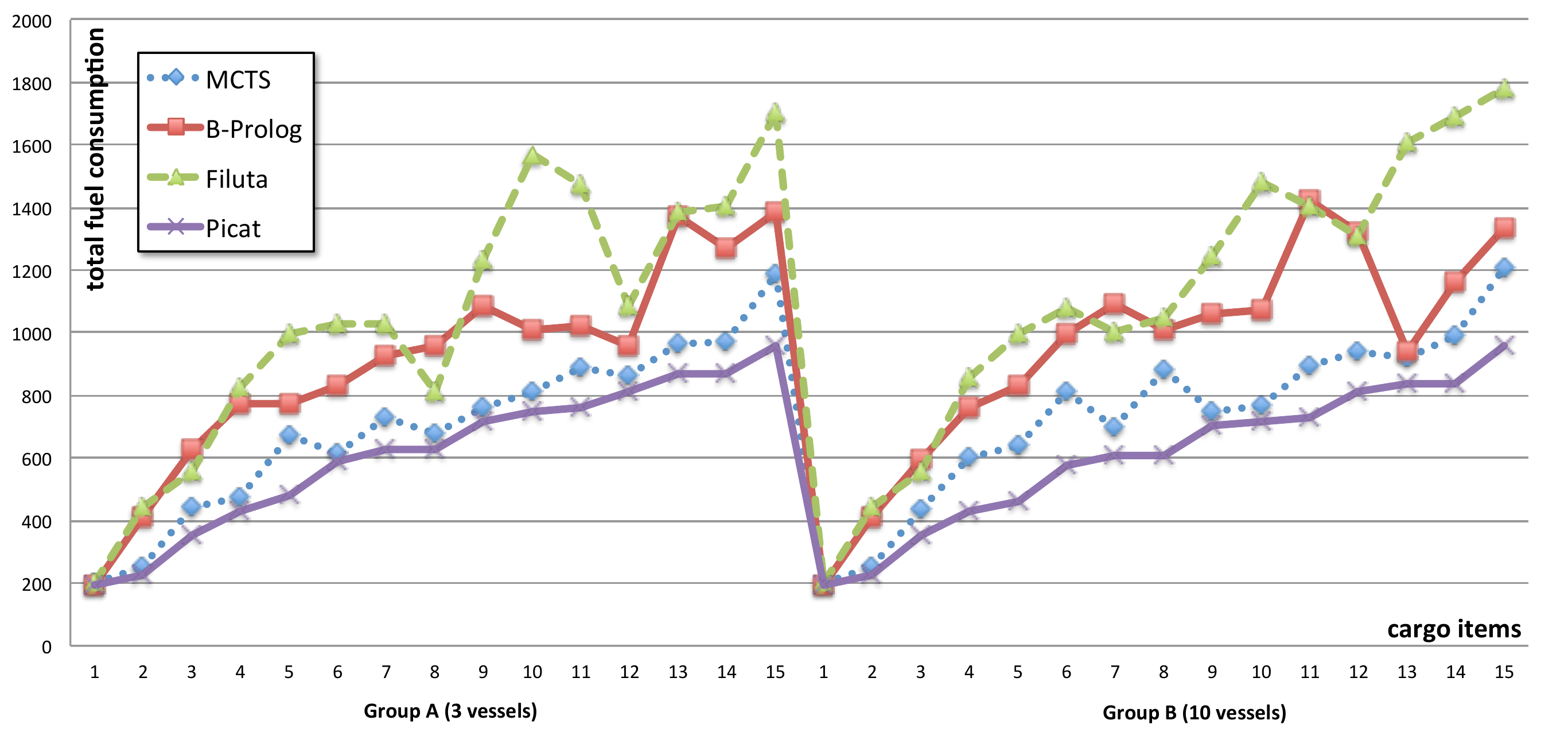}
   \includegraphics[width=0.7\textwidth]{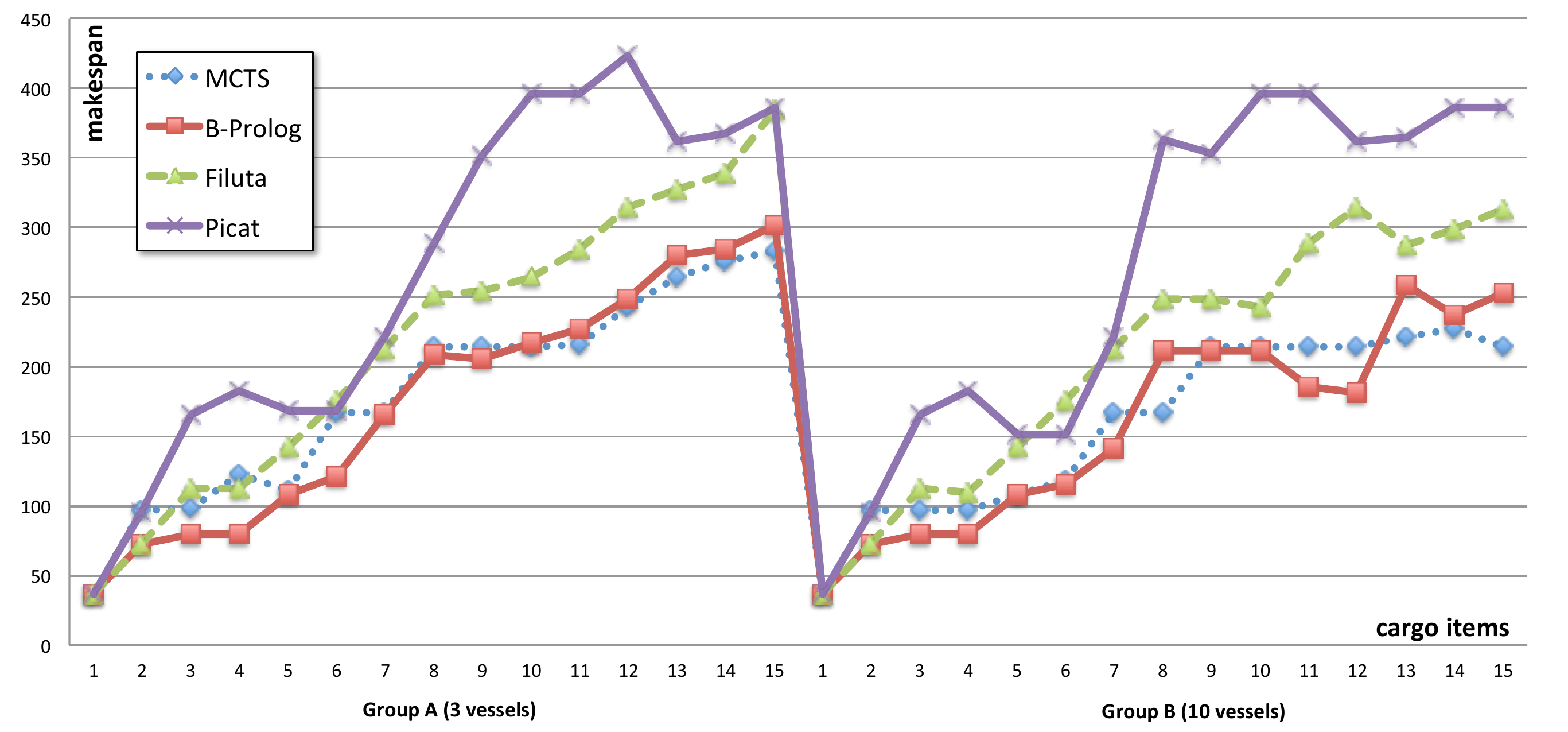} 
        \includegraphics[width=0.7\textwidth]{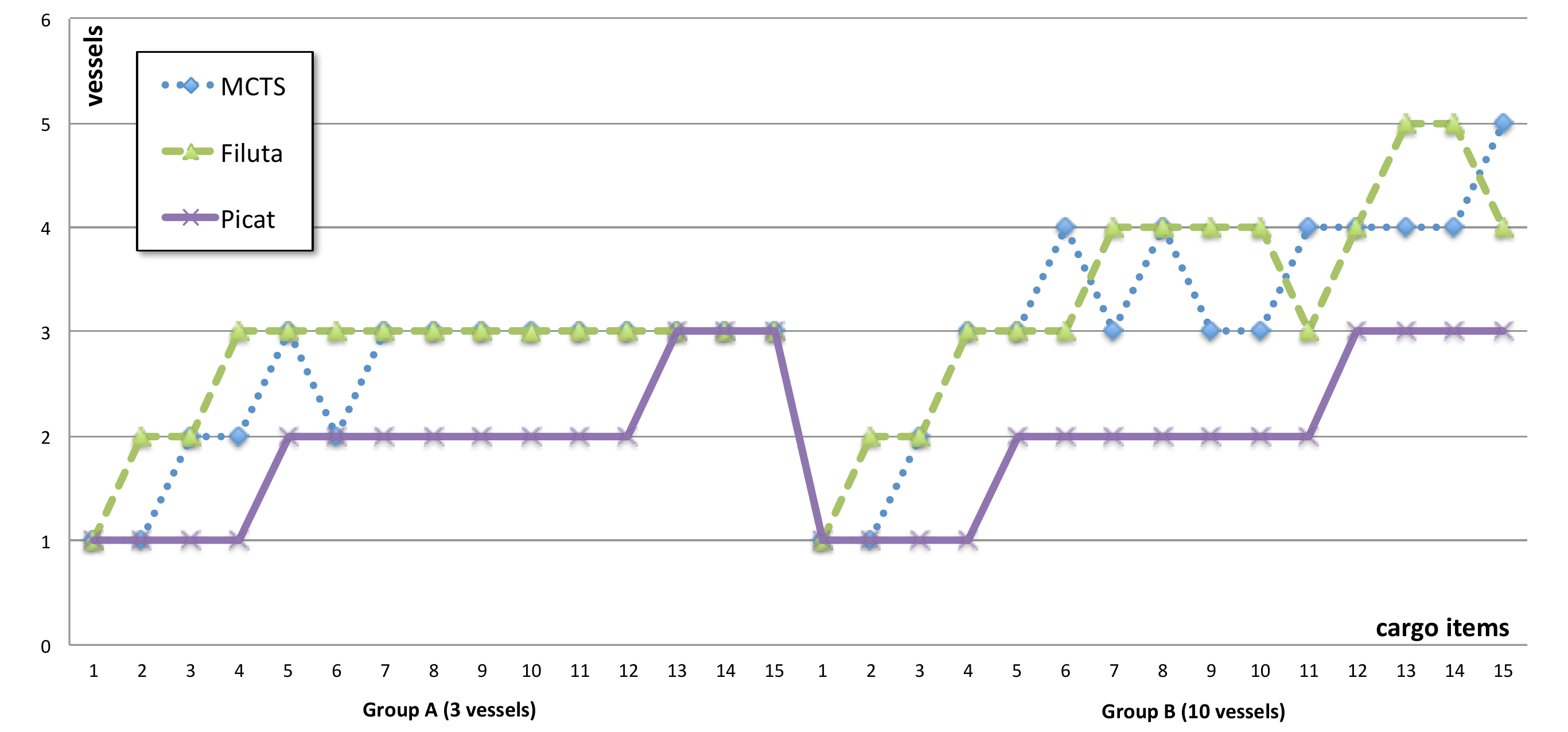}
 \end{center}
    \caption{Comparison of various metrics depending on the number of cargo items.}
    \label{GraphFuel}
 \end{figure}

\begin{figure}
 \begin{center}
   \includegraphics[width=0.8\textwidth]{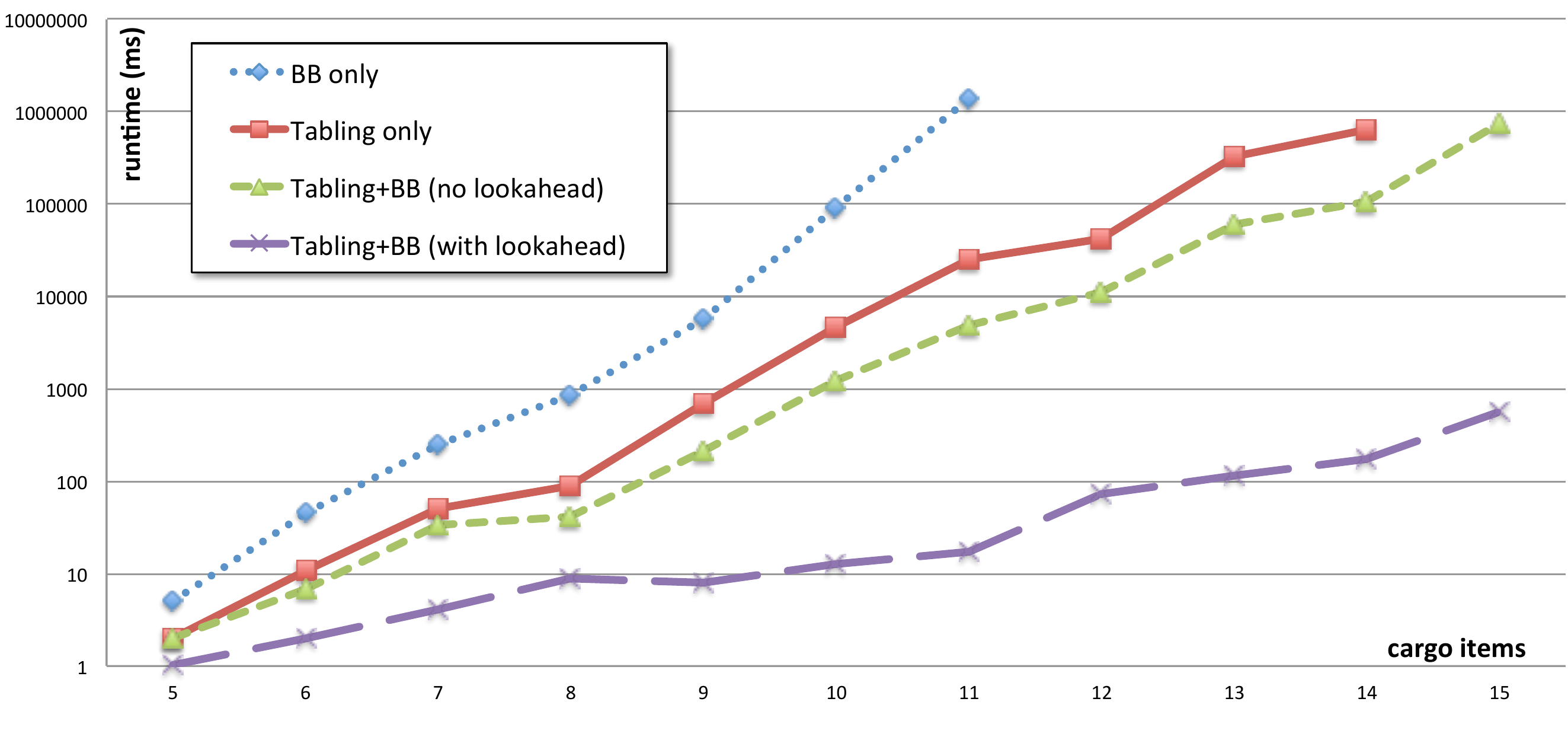}\\
 \end{center}
    \caption{Comparison of runtimes in milliseconds (logarithmic scale) for the pure tabling approach and for tabling enhanced by branch-and-bound.}
    \label{GraphRuntimes}
 \end{figure}

To understand better the behavior of the planners, we re-used the benchmark problems suggested in \cite{toropila2012}. These problems use the same network of waiting areas, ports, and platforms taken from the Petrobras problem. Two scenarios were proposed in \cite{toropila2012}, one with 3 vessels (Group A) and one with 10 vessels (Group B). For each scenario, we varied the number of cargo items from 1 to 15. Figure~\ref{GraphFuel} shows the comparison of various metrics for the discussed approaches (some data for the B-Prolog were not available). The experiment confirmed that the Picat-based planner finds the best plans regarding fuel consumption. Again, in all problems, the Picat-based planner run in less than one second (see Figure~\ref{GraphRuntimes}) and it proved optimality taking in account the assumptions restricting allowed plans. This comparison also clearly shows the effect of using more vessels in parallel. To achieve the smallest fuel consumption Picat usually used less vessels than Filuta and MCTS. When more vessels are used to deliver more cargo, the makespan goes down even if more cargo is to deliver. Docking cost is basically the same for all approaches as it is mainly determined by the time to load cargo and by the number of visits in ports (this cost depends on the time spent in ports).

To see the contribution of various Picat solving approaches we compared them using the benchmark set with 10 vessels. We used the Picat code as introduced in this paper, but we modified the solving mechanism. In particular, we compared the following approaches to solve the problem:
\begin{itemize}
\item using branch-and-bound with look-ahead but no tabling,
\item using tabling only without branch-and-bound,
\item using tabling with branch-and-bound but with no look-ahead
\item using tabling with branch-and-bound including look-ahead (this is the method that we used to compare with other approaches).
\end{itemize}
Figure~\ref{GraphRuntimes} shows the comparison of runtimes in milliseconds (a logarithmic scale). The conclusion from this experiment is obvious. Tabling is critical to solve the problem especially for route optimization as implemented using the procedure \verb(path_plan(. Without tabling, we cannot solve the problem with 12 cargo items within the limit of one hour. The reason is that path planning is done frequently in alternative top search branches (called in \verb(plan(), but with the same set of locations to visit. By using tabling the system stores the found optimal paths so the next time, it can re-use them rather than finding them again. Still, tabling alone is not enough and it runs our of memory for problems with 15 cargo items. Adding branch-and-bound is useful as it prevents exploring plans that are already worse than the best-so-far plan, but it decreases the runtime little. Branch-and-bound pays-off especially if we can estimate the ``future'' cost. Then the runtime is orders of magnitude smaller in comparison with individual solving techniques. In our case we estimated the fuel consumption to deliver not-yet delivered cargo.

\section{Conclusions}
In this paper we proposed to use tabling as the core technique for solving a real-life motivated Petrobras planning problem \cite{petrobras}. Proper representation of search states as well as the definition of the search space are important for tabling and so we discussed possible restrictions on the plans to be explored during search to make tabling a viable approach. We also implemented a form of resource-bounded search by adding ideas of branch-and-bound with simple look-ahead to tabling. We experimentally showed that the proposed method achieves best-so-far results regarding the quality of plans measured by total fuel consumption. Moreover, the runtime of the proposed planner is the lowest among all methods applied to this problem so far. The planner was implemented in Picat, but similar results can be achieved when using other Prolog-like languages with tabling such as B-Prolog.

We applied the ideas presented in this paper to other planning problems namely the Nomystery domain \cite{nomystery} and Sokoban  \cite{sokoban}. It seems that tabling is an appropriate method to solve such planning problems. Nevertheless, this paper also showed that tabling alone is not enough for optimization problems and that adding branch-and-bound with estimate of future cost is critical for runtime efficiency. We have implemented the branch-and-bound procedure aside tabling (using the blackboard mechanism) but the paper shows a direction for possible future improvements of the tabling mechanism to handle optimization problems better.

The Petrobras problem belongs among multi-criteria optimization problems. Optimizing fuel consumption seems to minimize the number of vessels used, but the negative consequence is increasing makespan. However adding time among the tabled arguments hampers performance as the B-Prolog planner showed \cite{bartak2013} and so the open question is how to handle such multi-criteria optimization. Though makespan is one of the metrics of the Petrobras problem, we understood that it is less important than fuel consumption and the number of vessels used. Practically, among the time-based criteria, it seems that delivering cargo by specified deadlines is more critical and this is the direction of our future extension towards a time-aware planner.

\section{Acknowledgements}
Roman Bart\'{a}k is supported by the Czech Science Foundation under the project PlanEx (P103-10-1287).

\bibliographystyle{acmtrans}
\bibliography{new_tlp2egui}

\label{lastpage}
\end{document}